\documentclass[letterpaper, 10 pt, conference]{ieeeconf}  

\IEEEoverridecommandlockouts                              

\usepackage{graphics} 
\usepackage{epsfig} 
\usepackage{times} 
\usepackage{amsmath} 
\usepackage{amssymb}  
 \usepackage{url}
\usepackage{multirow}
\usepackage{booktabs}
\usepackage{subcaption}
\usepackage{siunitx}
\title{\LARGE \bf
When Regression Meets Manifold Learning for Object Recognition and Pose Estimation
}

\author{Mai Bui$^{1}$, Sergey Zakharov$^{1,2}$, Shadi Albarqouni$^{1}$, Slobodan Ilic$^{1,2}$ and Nassir Navab$^{1,3}$
\thanks{$^{1}$ Department of Informatics, Technichal University Munich, Germany}%
\thanks{$^{2}$ Siemens AG, Munich, Germany}%
\thanks{$^{3}$ Whiting School of Engineering, Johns Hopkins University, USA}%
}

\begin{document}

\begin{minipage}{\textwidth}
	\large
	\textbf{IEEE Copyright Notice} \\
	\\
	$\copyright$
	2018 IEEE. Personal use of this material is permitted. Permission from IEEE must be obtained
	for  all  other  uses,  in  any  current  or  future  media,  including  reprinting/republishing  this
	material for advertising or promotional purposes, creating new collective works, for resale
	or redistribution to servers or lists, or reuse of any copyrighted component of this work in
	other works.\\
	\\
	Pre-print of article that will appear at the
	\textbf{2018 IEEE International Conference on Robotics and Automation (ICRA 2018).}\\
	\\
	\textbf{Please cite this paper as:}\\
	
	M. Bui, S. Zakharov, S. Albarqouni, S. Ilic and N. Navab, "When Regression Meets Manifold Learning for Object Recognition and Pose Estimation", IEEE International Conference on Robotics and Automation (ICRA)
	, 2018.\\

\end{minipage}

\newpage

\maketitle
\thispagestyle{empty}
\pagestyle{empty}

\begin{abstract}

In this work, we propose a method for object recognition and pose estimation from depth images using convolutional neural networks. Previous methods addressing this problem rely on manifold learning to learn low dimensional viewpoint descriptors and employ them in a nearest neighbor search on an estimated descriptor space. In comparison we create an efficient multi-task learning framework combining manifold descriptor learning and pose regression. By combining the strengths of manifold learning using triplet loss and pose regression, we could either estimate the pose directly reducing the complexity compared to NN search, or use learned descriptor for the NN descriptor matching. By in depth experimental evaluation of the novel loss function we observed that the view descriptors learned by the network are much more discriminative resulting in almost 30\% increase regarding relative pose accuracy compared to related works. On the other hand, regarding directly regressed poses we obtained important improvement compared to simple pose regression. By leveraging the advantages of both manifold learning and regression tasks, we are able to improve the current state-of-the-art for object recognition and pose retrieval that we demonstrate through in depth experimental evaluation.
\end{abstract}
\section{Introduction}

	3D object pose estimation and instance recognition is a widely researched topic in the field of computer vision with many application possibilities in augmented reality and robotics. Successfully recognizing a robot's surroundings and inferring object poses is crucial for tasks as robotic grasping. Here we offer a simple solution, the only requirement being a depth camera attached to the robot.
	
	
		\begin{figure}[h]

			\begin{center}
				\begin{subfigure}[b]{0.48\textwidth}
				\includegraphics[width=0.44\linewidth]{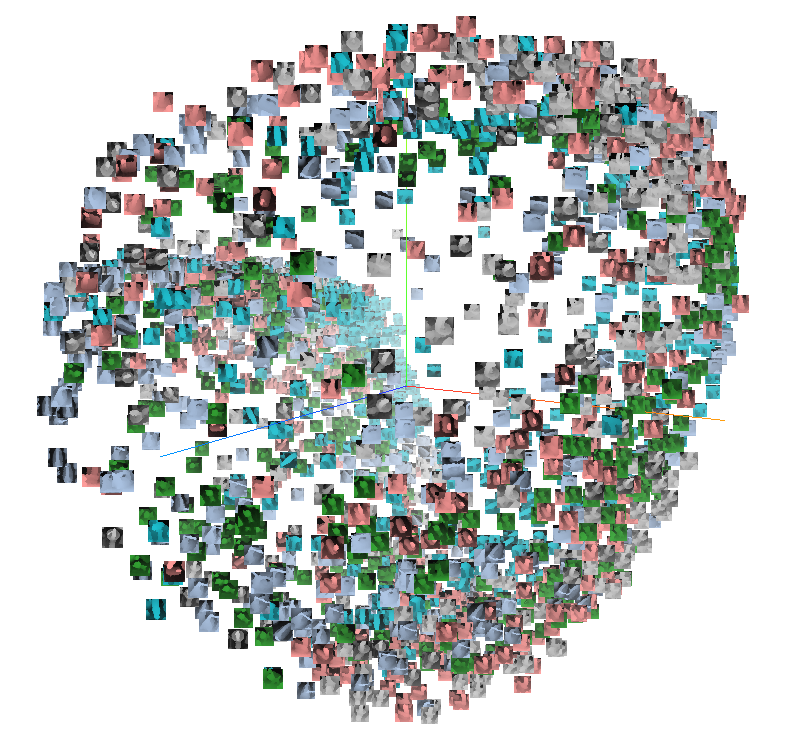}
				\includegraphics[width=0.44\linewidth]{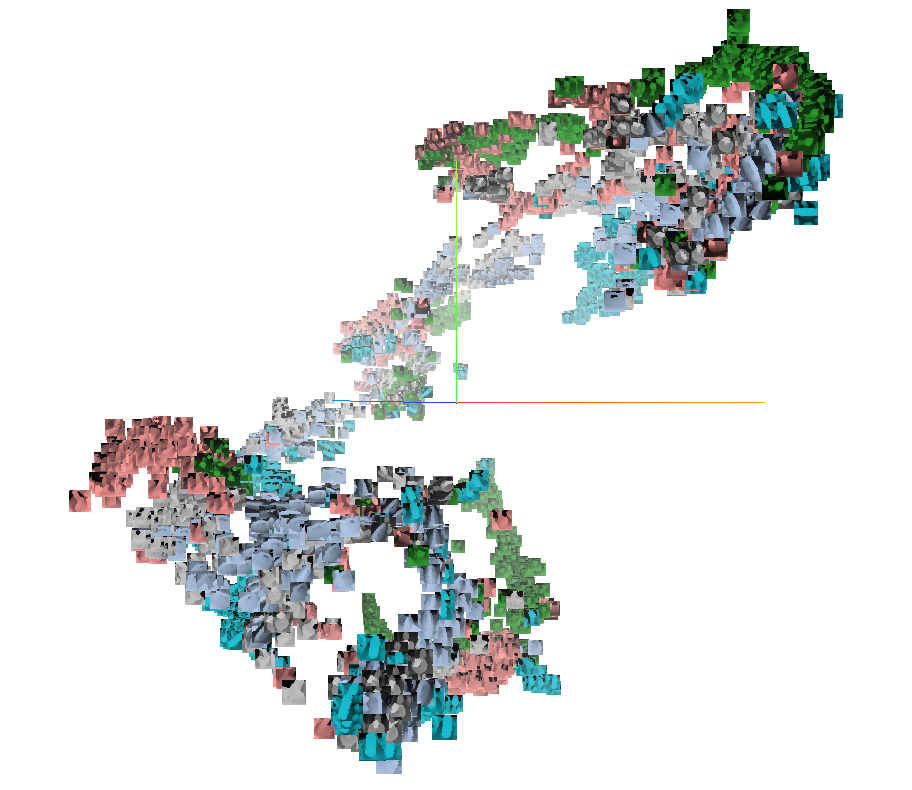}
				\caption{Obtained using direct pose regression.}
				\end{subfigure}
				\begin{subfigure}[b]{0.48\textwidth}
				\includegraphics[width=0.44\linewidth]{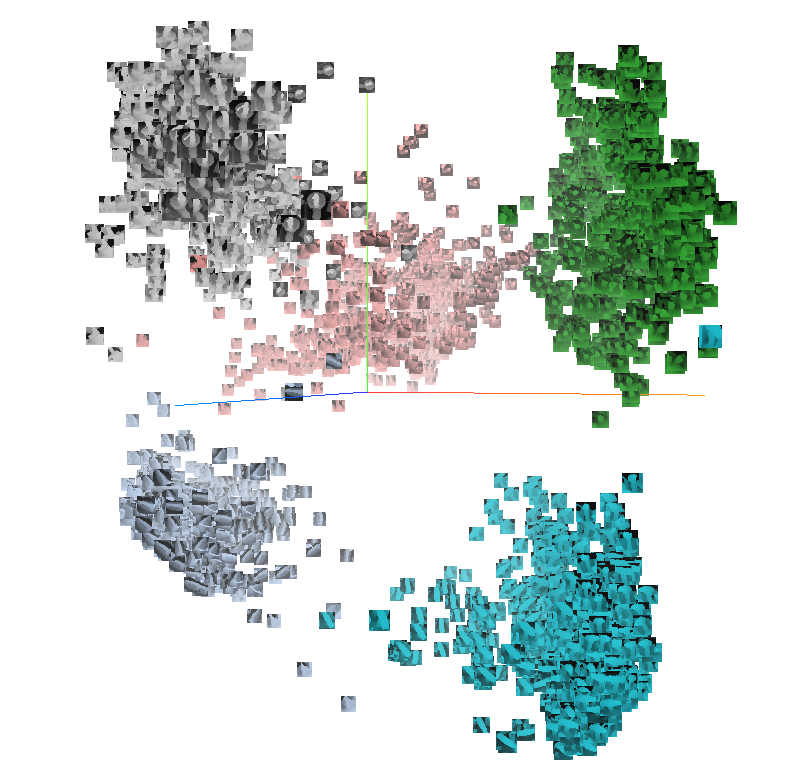}
				\includegraphics[width=0.44\linewidth]{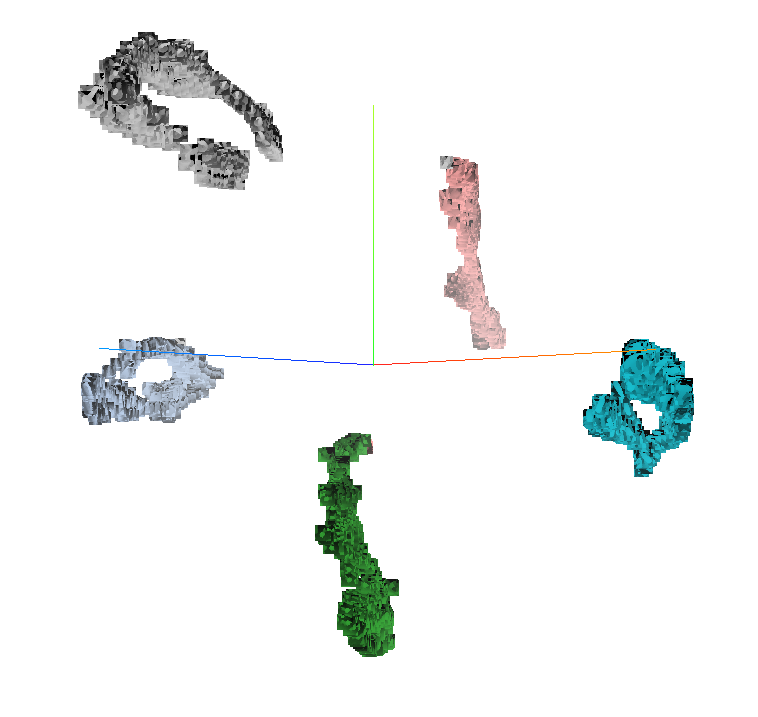}
				\caption{Obtained using our multi-task learning framework.}
				\end{subfigure}
				\caption{By using a multi-task learning framework, we are able to improve feature descriptors learned for object pose estimation. Depicted here is the feature visualization using left: PCA and right: t-SNE \cite{maaten2008visualizing} for five objects of the LineMOD \cite{hinterstoisser2012model} dataset.}
				\label{fig:fvisreg}
			\end{center}
			
		\end{figure}
	
	Even though deep learning has shown significant success in most computer vision tasks, its performance directly corresponds to the training set given and usually relies on very large datasets to perform and generalize well. However, there is an increasing availability of 3D models, especially in industrial applications, which can easily be used to create large amounts of synthetic images for training such models.
	Color or texture information might not always be available for CAD models and is difficult to include, while considering for example illumination changes as well. Realistic depth images are easier to obtain, and although they are usually noisy and can contain missing depth values, they incorporate more valuable information for object pose estimation and are, therefore, used as a main data source in our work. To this end, we rely on simulated depth renderings, assuming 3D models of the objects of interest are available.
	
	Previous methods working on object pose estimation have often used nearest neighbor search methods on hand-crafted \cite{hinterstoisser2012model} or learned feature representations \cite{kehl2016deep,wohlhart2015learning}. They describe the object seen from the particular viewpoint to retrieve the closest pose for a given test image and are able to predict poses with great accuracy. However, when including more objects, the complexity of those methods usually do not scale well. On the other hand, direct regression approaches offer great capabilities, but are not yet robust enough for real-world applications. 
	
	Inspired by the work first introduced by Wohlhart and Lepetit \cite{wohlhart2015learning}, we propose a multi-task learning pipeline, which combines the strengths of manifold learning and regression, to learn robust features from which the object's pose can be inferred. Fig. \ref{fig:fvisreg} depicts an example feature visualization obtained through regression and, in comparison to our proposed method, showing a significant improvement in discrimination in the feature space. Thus, we are able to combine the generalization capabilities shown in manifold learning tasks and the variability of regression into a deep learning framework for object pose estimation. To this purpose, we introduce a new loss function, which includes both manifold learning and regression terms. Therefore, we analyze how the two tasks influence each other and show that each task can be beneficial to one another in the context of estimating object poses.

	To summarize, our contributions described in this paper include the following:
	\begin{itemize}
		\item 
		introducing a loss function using a combination of regression and manifold learning to create an end-to-end framework for object recognition and pose estimation,
		\item
		improvement in accuracy and feature robustness compared to the current baseline methods, and
		\item 
		a detailed analysis and comparison of nearest neighbor pose retrieval and direct pose regression.
	\end{itemize}
	
	In the next sections, a detailed description of the proposed pipeline is given. Moreover, besides reporting the results of our method, a detailed analysis on comparing regression and nearest neighbor pose retrieval using feature descriptors is conducted. By using a combination of manifold learning and regression, we were able to significantly improve the regression performance as well as to create more robust feature descriptors compared to the baseline methods. Thus, we improved both aspects: nearest neighbor pose retrieval and direct pose regression with our framework and obtain a large accuracy boost compared to the related works.
	
\section{Related Work}
Detecting objects and estimating their 3D pose is a well researched topic in the field of computer vision \cite{tejani2014latent,schwarz2015rgb,hodavn2015detection,brachmann2016uncertainty,brachmann2014learning,krull2015learning}. To explain a few, Brachmann et al. \cite{brachmann2016uncertainty} use random forests to predict object labels as well as object coordinates from which the pose can then be inferred. In \cite{brachmann2014learning} the authors extend their approach by using an auto-context framework and exploiting the uncertainty over object labels and coordinate probabilities to improve their method. Whereas Krull et al. \cite{krull2015learning} use a convolutional neural network (CNN) on top to score estimated object probabilities, coordinates and depth images compared to the ground truth and optimize accordingly. 
Other approaches have proposed to use 2D view-specific templates for object detection and pose estimation. By computing handcrafted feature representations for a known set of views, most similar matches can easily be found for a given template to infer its class and pose \cite{hinterstoisser2011multimodal}.
To improve this pipeline several approaches have proposed to use learning-based methods, instead of relying on handcrafted features, to infer more descriptive and robust feature representations for object pose retrieval. Therefore, as those are most related to our work, we will first describe some descriptor learning approaches, before introducing methods working on direct pose regression.\\
 
\textbf{Descriptor Learning.} For instance, Kehl et al. \cite{kehl2016deep} use convolutional auto-encoders to learn feature descriptors from RGB-D image patches. Zamir et al. \cite{zamir2016generic} use a siamese network architecture to compute features based on object-centric viewpoint matches for camera pose estimation. Further, they show that the resulting model generalizes well to other tasks, including object pose estimation. Still, this method as is has not been shown to generalize to multiple objects in a cluttered environment.

Wohlhart and Lepetit \cite{wohlhart2015learning} propose a descriptor learning approach using CNNs, which is most related to our work. By enforcing the Euclidean loss between images from similar views to be close and from different objects to be far away both the object's identity and pose information can be stored in highly separable feature descriptors. The pose for a given test image can then be estimated by nearest neighbor lookup to find the closest corresponding pose of the found object. One of the main drawbacks of this method is that in-plane rotations are not considered by the pipeline, which is rarely the case in real-world applications. 

Zakharov et al. \cite{zakharov2017} include in-plane rotations to the above mentioned method and improve it by introducing an updated triplet loss function in which the margin term value is set to be dynamic depending on the type of the negative sample, as opposed to the former method. However, this method, as well as the former one, relies on nearest neighbor search, the complexity of which grows with respect to the number of objects. 

\textbf{Pose Estimation.} Several state-of-the-art methods on object pose estimation propose using pixel to 3D point correspondence prediction based on random forest and iterative pose refinement using RANSAC \cite{shotton2013scene}. Due to the recent successes of this work, it has been extended and optimized in several related methods \cite{guzman2014multi, valentin2015exploiting}.

In comparison, more recent methods have introduced direct regression approaches, e.g. PoseNet \cite{kendall2015posenet}, where a CNN is employed to regress the position and orientation of a camera given an RGB image. While this method is able to infer the camera's six degrees of freedom (DoF) in an end-to-end fashion using only an RGB image as input, the reported accuracy is still significantly lower than the reported results based on point correspondence prediction.


Based on the analysis of previous methods, we propose to combine the strengths of both regression and manifold learning to obtain separable feature descriptors and to leverage the advantages of both methods for the problem of object recognition and pose estimation.

%
%
%
%

\section{Methodology}

\begin{figure*}[h]
	\begin{center}
		\includegraphics[width=0.95\linewidth]{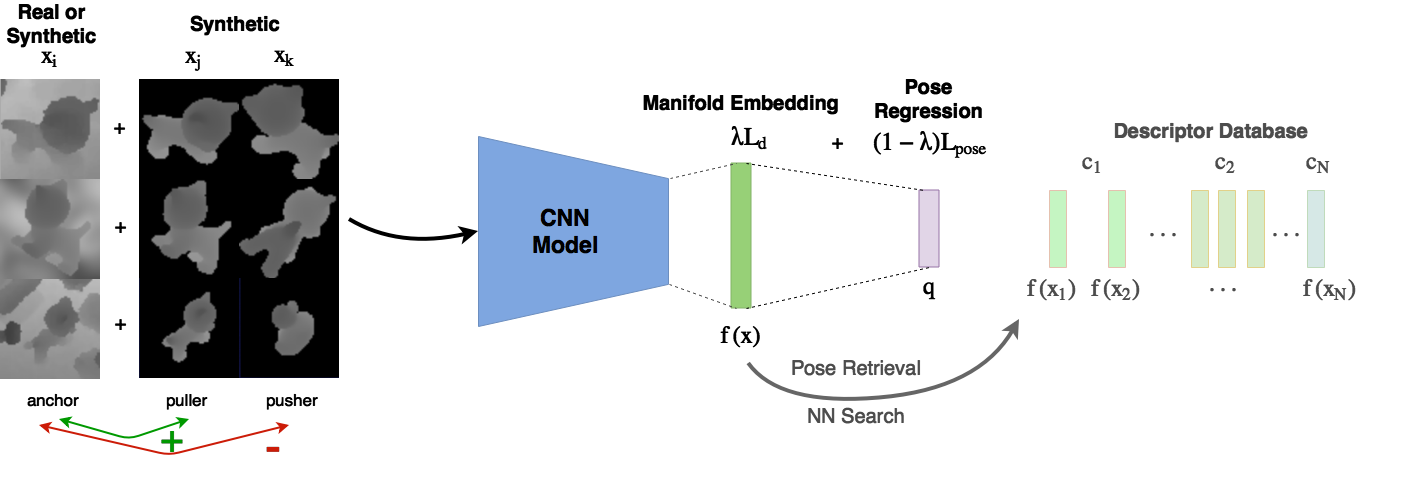}
		\caption{Given an input depth image patch $\textbf{x}_i$, we create corresponding triplets ($\textbf{x}_i, \textbf{x}_j, \textbf{x}_k$) and pairs ($\textbf{x}_i, \textbf{x}_j$) to optimize our model on both manifold embedding, creating robust feature descriptors, and pose regression. Obtaining either a direct pose estimate $\textbf{q}$ or using the resulting feature descriptor for nearest neighbor search in the descriptor database.}
		\label{fig:pipeline}
	\end{center}
\end{figure*}
Our methodology, depicted in Fig. \ref{fig:pipeline}, starts with training a CNN model for a given training set $S_{train} = \{ \textbf{s}_1, \ldots, \textbf{s}_N \} = \{(\textbf{x}_1, {c}_1, \textbf{q}_1), \ldots, (\textbf{x}_N, {c}_N ,\textbf{q}_N)\}$ consisting of $N$ samples. Each sample $\textbf{s}$ comprises a depth image patch $\textbf{x} \in \mathbb{R^{\text{n} \times \text{n}}}$ of an object, identified by its class $c \in \mathbb{N}$, together with the corresponding pose vector, $\textbf{q} \in \mathbb{R^\text{4}}$, which gives the orientation represented by quaternions.

Our objective is to model the mapping function $\phi: X \rightarrow Q$, thus for a given input $\textbf{x}$ the predicted pose vector $\hat{\textbf{q}}$ is obtained as 

\begin{equation}
	\hat{\textbf{q}} = \phi(\textbf{x}; \textbf{w}),
\end{equation}
where $\textbf{w}$ are the model parameters. While the primary objective is to obtain an accurate pose estimation for any unseen data, having a well clustered feature space is of high interest as well and can be used to identify the objects class, if needed. To achieve this, we model the problem as a multi-task learning, namely pose regression and descriptor learning. Thus the overall objective function can be written as

\begin{equation}
	L_{MTL} = (1 - \lambda) L_{pose} + \lambda L_{d},
\end{equation}
where $\lambda$ is a regularization parameter. $L_{pose}$ and $L_d$ are the objective functions for the pose regression task and the descriptor learning task respectively.
\subsection{Regression}

During training, our CNN model maps a given input $\textbf{x}$ to a lower dimensional feature vector $f(\textbf{x}) \in \mathbb{R}^d$, i.e. the output of the last fully connected layer before it is further utilized to regress the pose using the following loss function:
\begin{equation}
L_{pose} = \|\textbf{q} - \frac{\hat{\textbf{q}}}{\| \hat{\textbf{q}} \|} \|_2^2,
\end{equation}
where $\| \cdot \|_2$ is the $l_2$-norm and $\textbf{q}$ is the corresponding ground truth pose.

%
%
%

\subsection{Descriptor Learning}
To create robust feature descriptors, object identities as well as poses should be well differentiable in the feature space, creating a compact clustering of object classes as well as a respective pose mapping within the clusters. As a second requirement, since we mainly train our model on synthetic images, we need to map synthetic and real images to the same domain to ensure generalization to real applications. Here, we use the triplet and pairwise loss, as introduced in \cite{wohlhart2015learning}. Overall, we obtain the following loss function $L_{d}$ for descriptor learning:
\begin{equation}
L_{d} = L_{triplets} + L_{pairs}.
\end{equation}
As depicted in Fig. \ref{fig:pipeline}, our model is trained on a set of triplets $(\textbf{s}_i, \textbf{s}_j, \textbf{s}_k) \in T$, where sample $\textbf{s}_i$ (anchor) corresponds to the current image $\textbf{x}_i$  and $\textbf{s}_j$ (puller) is chosen so that the image corresponds to the same object $c_i$ viewed from a similar pose $\textbf{q}_j$. However, $\textbf{s}_k$ (pusher) is chosen so that the image $\textbf{x}_k$ corresponds either to a different object $c_k$ or the same object $c_i$, but viewed under a very different pose $\textbf{q}_k$. 
The resulting loss, $L_{triplets}$, defined over a batch of triplets is formulated as
\begin{equation}
	L_{triplets} = \mkern-18mu \sum_{(\textbf{s}_i,\textbf{s}_j,\textbf{s}_k) \in T} { \mkern-18mu max\left(0,1-\frac{||f(\textbf{x}_i)-f(\textbf{x}_k)||_2^2}{||f(\textbf{x}_i)-f(\textbf{x}_j)||_2^2+m}\right)},
\end{equation}
pulling viewpoints under similar poses close together and pushing dissimilar ones or different objects further away. As appeared in \cite{zakharov2017}, $m$ corresponds to a dynamic margin defined as:
\begin{equation}
	m = \begin{cases}
	2 \arccos (|\textbf{q}_i \cdot \textbf{q}_j|) &\text{if } c_i = c_j,\\
	\gamma &\text{else,}
	\end{cases}
\end{equation}
where  $\gamma > 2\pi$.
The dynamic margin ensures that objects of different classes get pushed farther away while the margin for the same objects depends on the angular distance between the current viewpoints $\textbf{q}_i$ and $\textbf{q}_j$.

In addition, the pair-wise loss $L_{pairs}$ is used to push together the sample feature descriptors of the same object under the same or very similar pose but with different backgrounds or coming from different domains (synthetic and real). The pair-wise loss is computed on pairs $(\textbf{s}_i, \textbf{s}_j) \in P$ and is defined as:
\begin{equation}
L_{pairs} = \sum_{(\textbf{s}_i,\textbf{s}_j) \in P} {||f(\textbf{x}_i)-f(\textbf{x}_j)||_2^2},
\end{equation}
$f(\textbf{x}_i)$ being the feature descriptor extracted from the neural network for image $\textbf{x}_i$.



\section{Experimental Setup}
In this section, we first describe how the dataset is generated by combining simulated object renderings with background noise and real images taken from the LineMOD dataset \cite{hinterstoisser2012model}. Then we give an overview of our experimental setup before demonstrating and evaluating the results.
\subsection{Dataset Generation}
Since both former closely related feature descriptor learning and pose estimation methods used the LineMOD dataset for their experiments, we also chose this dataset to be able to evaluate the algorithm's performance. 

The LineMOD dataset consists of fifteen distinct 3D mesh models and respective RGB-D sequences of them together with the camera poses. These data are used to create training set $S_{train}$, database set $S_{db}$, and test set $S_{test}$ each consisting of samples $\textbf{s} = (\textbf{x}, c, \textbf{q})$, where $\textbf{x}$ stands for the image patch, $c$ and $\textbf{q}$ are the corresponding class and pose, respectively. The training set $S_{train}$ as its name suggests used exclusively for training. The database set $S_{db}$ is used for the evaluation phase, where its samples are used to construct a descriptor database used for the nearest neighbor search, whereas $S_{test}$ is used exclusively in the evaluation phase.

\begin{figure}[!htbp]
	\centering
	\begin{subfigure}[b]{.32\linewidth}
		\centering
		\includegraphics[width=\linewidth]{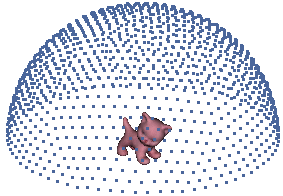}
		\caption{Regular}
		\label{fig:normal}
	\end{subfigure} \hfill
	\begin{subfigure}[b]{.32\linewidth}
		\centering
		\includegraphics[width=\linewidth]{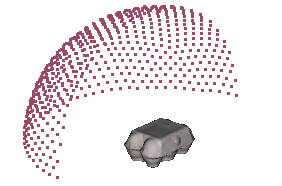}
		\caption{Symmetric}
		\label{fig:symmetric}
	\end{subfigure} \hfill
	\begin{subfigure}[b]{.32\linewidth}
		\centering
		\includegraphics[width=\linewidth]{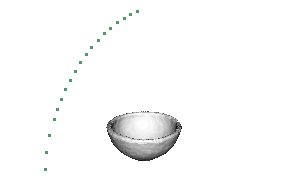}
		\caption{Rotation-inv.}
		\label{fig:rotinv}
	\end{subfigure}
	\caption{Sampling points for different objects types: vertices represent camera positions from which the object is rendered.}
	\label{fig:samplings}
\end{figure}

First, we render each of the fifteen objects from different viewpoints covering their upper hemisphere, depicted in Fig. \ref{fig:normal}. Here, following previous methods \cite{wohlhart2015learning,zakharov2017}, viewpoints are defined as vertices of an icosahedron centered around the object. By repeatedly subdividing each triangular face of the icosahedron additional viewpoints and, therefore, a denser representation can be created. In our case, the training set $S_{train}$ sampling is achieved by recursively applying three consecutive subdivisions on the initial icosahedron structure (see Fig. \ref{fig:normal}). Furthermore, following the method of \cite{zakharov2017}, we add in-plane rotations at each vertex position by rotating the camera at each sampling vertex from -45 to 45 degrees using a stride of 15 degrees.

As a next step, we extract patches covering the objects from both rendered images and real images coming from the depth sequences. The bounding box is defined by the bounding cube size of 40 \si{cm^{3}} centered on the object. All the values beyond the bounding cube are clipped. Upon extraction of the patches, each of them is being normalized, mapped to the range of $[0, 1]$ and stored together with its identity class $c$ and pose $\textbf{q}$ resulting in a single sample $\textbf{s}$.

The training set $S_{train}$ is then generated by combining all the samples coming from the rendering and $50\%$ of the real data from the depth sequences (resulting in approximately $18\%$ of the real data in $S_{train}$). The selection of the real samples is performed by choosing the most similar poses to the ones used for synthetically rendered samples. The rest of the real samples are used to generate the test set $S_{test}$. The database set $S_{db}$ contains only the synthetic part of the $S_{train}$.

\subsubsection{Treating Rotation-invariant Objects}
Four out of fifteen objects of the LineMOD dataset have a property of rotation-invariance and introduce an ambiguity to the generation of triplets needed for the triplet classification loss. For instance, the \textit{bowl} object is fully rotation-invariant, whereas the \textit{cup}, \textit{eggbox} and \textit{glue} object are only rotation-invariant around a single plane or, in other words, symmetric. These four objects need to be treated differently from the rest. This comes from the fact that, in the case of rotation-invariant objects, samples representing different poses might look exactly the same, which can result in a faulty triplet required for the triplet loss function. 

To solve this problem, we render only a limited amount of poses for those objects, such that every image patch is unique. Sample vertices for different object types are demonstrated in Fig. \ref{fig:symmetric} and \ref{fig:rotinv}. Since both training set $S_{train}$, and test set $S_{test}$ also include real samples we also omit ambiguous poses in them and only consider those that are close to the ones coming from the renderer. Note that this sampling also results in an unbalanced percentage of real images included for each object. To consider this, we create datasets of five, ten and fifteen objects and only include the rotation-invariant objects when using the full LineMOD dataset of fifteen objects.

\subsubsection{Data Augmentation}
The synthetic samples coming from the renderer have a black background, which makes them very different from the real samples. Since we have a limited amount of the real data available and cannot cover all the possible poses, we augment the training samples with a background noise whenever the real sample for this pose is not available. Augmented samples are included in the training set $S_{train}$ during training in an online fashion, generating different noise patterns for each anchor sample. The noise type we use for our pipeline is purely synthetic and is present in both \cite{wohlhart2015learning} and \cite{zakharov2017} showing the best performance among synthetic noise types. It is referred to as fractal noise and is based on a combination of several octaves of simplex noise first introduced in \cite{perlin2001noise}. It provides a smooth non-interrupting noise pattern and is often used for landscape generation by game developers. 


\subsection{Implementation Details}
With this training and testing setup, we extract patches of size $n=64$. We then train our CNN, where, if not stated otherwise, we use the network architecture introduced in \cite{wohlhart2015learning}, except we set the feature descriptor size to $d=64$. As it is mentioned in \cite{wohlhart2015learning} at some point increasing the feature descriptor size does not improve the methods performance anymore. As for regression we found a similar effect during our experiments, in which we experienced $d=64$ to be a good trade-off between nearest neighbor and regression performance. 
During our experiments, we set $\gamma=10.0$ for the dynamic margin. 
The network was trained on a Linux-based system with 64GB RAM and 8GB NVIDIA GeForce GTX 1080 graphics card. All experiments are implemented using TensorFlow\footnote{https://www.tensorflow.org/} with Adam optimizer and an initial learning rate of $1e^{-3}$, while the batch size was set to $300$.

\subsection{Baseline Models}
To analyze not only our method, but the effect of multi-task learning, i.e. regression and learning robust feature descriptors together, we report the results compared to the baseline method \cite{zakharov2017}. Here we train on the loss function $L_d$ to compare to the results obtained by nearest neighbor pose retrieval, abbreviated as \textit{NN}. However, in comparison, the baseline in their original work uses RGB-D data and includes normals as additional information whereas we only use depth information. Nevertheless, the authors provided us with an implementation of their pipeline written in python so that we were able to run all our experiments and compare the results using depth images only. 
In addition, we conduct our own baseline for regression only ($R$) and report the results. Furthermore, to evaluate our method, we report the results obtained by the end-to-end regression (\textit{Rours}), as well as the results obtained by using the resulting features for nearest neighbor lookup (\textit{NNours}). 

\subsection{Evaluation Metrics}
To evaluate the performance of our method, we use the angular error comparing the ground truth pose $\textbf{q}$ and the predicted one $\hat{\textbf{q}}$ for a given test image $\textbf{x}_i$:

\begin{equation}
\theta(\textbf{q}_i, \hat{\textbf{q}_i}) = 2 \cdot arccos(|\textbf{q}_i \cdot \hat{\textbf{q}_i}|).
\end{equation}
The pose can either be obtained by nearest neighbor lookup or directly estimated by the neural network.
Furthermore, we report the angular accuracy, where each test image is considered a true positive if its angular error is below a threshold $t$, where $t \in [10, 20, 40]$ degrees.

Also, to evaluate the resulting feature descriptor we visualize the features in a lower dimensional space using PCA and t-SNE \cite{maaten2008visualizing}.

\section{Evaluation}
In this section, we first give a detailed analysis of our method compared to the baseline and discuss several aspects of our method, starting with the influence of the network architecture on the methods performance.
\subsection{Comparison to baseline method}
We first evaluated our method on models trained on a different number of objects, for which the mean angular error is reported in Table \ref{tab:results}. Note that for nearest neighbor pose retrieval only the poses of correctly classified objects are considered, while for regression the whole test set is used, as in this case we do not directly infer the class. 

Overall, we experienced a significant improvement in performance for both regression as well as nearest neighbor search accuracy by our proposed method. During training, the usually more difficult regression task, seems to be optimized by additionally focusing on learning a meaningful embedding, improving the mean angular error by $28.8\%$. Since poses as well as objects are already well-distinguished and the feature descriptors separated by the triplets and pair loss functions, regression can more easily be learned. 

As for the performance of nearest neighbor search we found an improvement in robustness and accuracy of our multi-task learning framework compared to the baseline. The standard deviation as well as the mean angular error of our model, \textit{NNours}, decreases significantly, making the method more robust. Here we can report a relative improvement of $30.0\%$ for the mean angular error while training on the full LineMOD dataset, meaning fifteen objects.

Both regression and nearest neighbor method benefit from jointly learning robust features and poses. Which model to choose now becomes a trade-off between time complexity and accuracy, which we will address further in section \ref{sec:scalability}.
\begin{table*}[h]
	\centering
	\resizebox{\textwidth}{!}{%
		\begin{tabular}{@{}lcccccc@{}}
			\toprule
			& \multicolumn{2}{c}{\textbf{15 Objects}}                 & \multicolumn{2}{c}{\textbf{10 Objects}}                 & \multicolumn{2}{c}{\textbf{5 Objects}}                  \\ \cmidrule(l){2-3} \cmidrule(l){4-5} \cmidrule(l){6-7}
			& \textbf{Mean (Median) $\pm$ Std} & \textbf{Classification} & \textbf{Mean (Median) $\pm$ Std} & \textbf{Classification} & \textbf{Mean (Median) $\pm$ Std} & \textbf{Classification} \\ \midrule
			\textbf{\textit{NN} \cite{zakharov2017}}     & 25.29$^{\circ}$ (11.76$^{\circ}$) $\pm$ 40.75$^{\circ}$           & 92.46\%                   & 19.98$^{\circ}$ (\textbf{10.58}$^{\circ}$) $\pm$ 34.78$^{\circ}$           & 92.56\%                   & 24.19$^{\circ}$ (10.72$^{\circ}$) $\pm$ 43.34$^{\circ}$           & 99.31\%                   \\
			\textbf{\textit{NNours}} & \textbf{17.70$^{\circ}$ (11.59$^{\circ}$) $\pm$ 25.78$^{\circ}$}           & \textbf{97.07\%}                   & \textbf{14.74$^{\circ}$} (11.53$^{\circ}$) $\pm$ \textbf{15.04$^{\circ}$}           & \textbf{97.50\%}                   & \textbf{13.05$^{\circ}$ (10.29$^{\circ}$) $\pm$ 15.19$^{\circ}$}           & \textbf{99.90\%}                   \\ \midrule
			\textbf{\textit{R}}      & 38.23$^{\circ}$ (26.16$^{\circ}$) $\pm$ 34.65$^{\circ}$           & -                       & 29.17$^{\circ}$ (20.69$^{\circ}$) $\pm$ 28.03$^{\circ}$           & -                       & 22.07$^{\circ}$ (15.56$^{\circ}$) $\pm$ 24.40$^{\circ}$           & -                       \\
			\textbf{\textit{Rours}}  & \textbf{27.28$^{\circ}$ (19.25$^{\circ}$)} $\pm$ \textbf{27.26$^{\circ}$}           & -                       & \textbf{23.08$^{\circ}$ (17.56$^{\circ}$)} $\pm$ \textbf{21.25$^{\circ}$}           & -                       & \textbf{19.16$^{\circ}$ (13.80$^{\circ}$)} $\pm$ \textbf{21.54$^{\circ}$}           & -\\              
			\bottomrule
		\end{tabular}
	}
	\caption{Angular error of the baseline method (\textit{NN}), regression (\textit{R}) and our approach (\textit{Rours, NNours}).}
	\label{tab:results}
	\vspace{-5pt}
\end{table*}

Next, to analyze our resulting feature descriptors, we compare our method using nearest neighbor search, \textit{NNours} to the baseline method, for which we report the classification and pose accuracy in Table \ref{tab:percent}. Again, our experiment results show that the feature descriptors provided by the model trained on both tasks seem to be more differentiable as well. As a result the nearest neighbor pose retrieval accuracy improves even further. Additionally, we are able to improve the classification accuracy compared to the state-of-the art methods.

%

\begin{table}[h]
	\centering
	\resizebox{1\linewidth}{!}{%
		\def\arraystretch{1}%
		\begin{tabular}{@{}lcccc@{}}
			\toprule
			\multicolumn{1}{c}{\multirow{2}{*}{}} & \multicolumn{3}{c}{\textbf{Angular error}} & \multirow{2}{*}{\textbf{Classification}} \\ \cmidrule(lr){2-4}
			\multicolumn{1}{c}{} &                 \textbf{10$^{\circ}$} & \textbf{20$^{\circ}$} & \textbf{40$^{\circ}$} &   \\
			\midrule
			\textbf{\textit{NN}} \cite{zakharov2017} & 35.98\% &  71.56\% & 82.72\% & 92.46\% \\
			\textbf{\textit{NNours}} & 37.89\% &  79.61\% & 92.27\% & 97.07\% \\
			\textbf{\textit{NNours$_{deeper}$}} & \textbf{41.32\%} & \textbf{82.52\%} &  \textbf{93.51\%} & \textbf{97.26\%}\\ \bottomrule
		\end{tabular}
	}
	\vspace{0.4cm}
	\caption{Comparison between the classification and angular accuracy of the baseline method, \textit{NN}, and our results on 15 objects of the LineMod dataset.}
	\label{tab:percent}    
\end{table}

\subsection{Influence of Network Architecture}
Additionally, to explore the methods performance using network architectures with varying depths, we run our model using the network architecture described in \cite{bui2017x}. This architecture is two layers deeper and removes max pooling layers by including convolutional layers with stride two. Stated by the authors of \cite{wohlhart2015learning}, a deeper network architecture did not seem to improve the accuracy of the method further, which we experienced in our test as well, however by using our multi-task learning framework and testing on a deeper network architecture we found that we can improve the pose estimation accuracy even further. Here we were able to achieve the results seen in Table \ref{tab:percent}, abbreviated as {\textit{NNours$_{deeper}$}. We report a relative improvement of $7.2\%$ using nearest neighbor search and $9.0\%$ in the mean angular error of our regression results by using a deeper network architecture, while training on the full LineMOD dataset. We believe that by optimizing the network further, we can achieve even better regression accuracy.

	\subsection{Feature Visualization}
	As we have shown in our experiments, apart from the improvement in accuracy for pose regression, we experienced an increase in the performance of nearest neighbor pose retrieval as well. To analyze the resulting feature descriptors, we visualize the descriptors in the lower dimensional 3D-space using PCA and t-SNE. 
	
	
	We use TensorBoard\footnote{https://www.tensorflow.org/get\_started/embedding\_viz} to create the visualizations. For t-SNE, we use a perplexity of $100$, learning rate of $10$ until convergence. Using PCA, the variance including the best three components resulted in $53.2\%$. The resulting clusters for five objects can be seen in Fig. \ref{fig:fvisreg}. We observe that in both cases the object classes are nicely distinguished using feature descriptors obtained by our proposed approach. 
	
	\subsection{Scalability}
	\label{sec:scalability}
	In this section we analyze the time complexity and accuracy of our models at different number of objects. For nearest neighbor search we use a standard OpenCV matcher. Fig. \ref{fig:scalability} shows the mean time of our models and the corresponding angular error. The mean time for regression is calculated as one forward pass of the neural network. For nearest neighbor methods only the matching time is tracked. To obtain the total time needed, the time of one forward pass should be added to the shown results for matching. One can see nicely that regression has a constant time, regardless of how many objects are used, whereas for nearest neighbor searches the time increases with additional objects. Depending on the application, this and the drop in accuracy for additional objects should be taken into account.
		\begin{figure}[b]
			\begin{center}
				\includegraphics[width=0.97\linewidth]{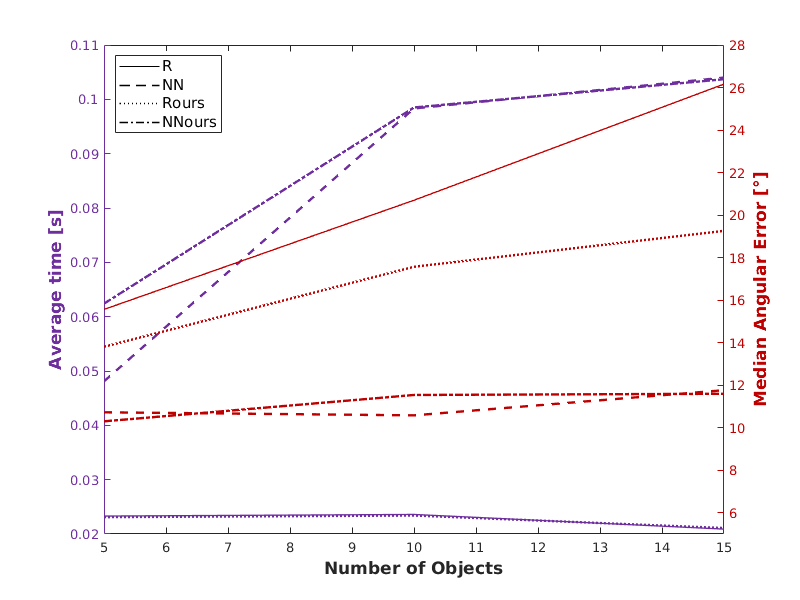}
				\caption{Average time and median angular error of nearest neighbor pose retrieval, regression and our approach.}
				\label{fig:scalability}
			\end{center}
		\end{figure}

	\subsection{Sensitivity to regularization parameter $\lambda$}
	Since our loss function includes a regularization parameter $\lambda$ balancing the two components of regression and manifold learning, we conducted experiments on the sensitivity of this parameter using the full LineMOD dataset. By choosing different values for $\lambda$ and thus weighting either the $L_{d}$ loss or the pose loss $L_{pose}$ more, we found that the results improve for nearest neighbor pose retrieval, if the two terms are equally weighted, and decreases when focusing more on the regression loss. Regarding regression, we observed similar results: improvement when additionally focusing on the $L_{d}$ loss, enhancing the feature representation and decrease, if the model is only trained on regression. Nevertheless, it can be seen that regression has a much stronger influence on the nearest neighbor pose retrieval in terms of performance than the other way around.
	
	The results, depicted in Fig. \ref{fig:lambda}, emphasize our assumption that the two terms are beneficial to one another, i.e. both features and pose regression are mutually optimized. Note that we omit the regression result in case the model was only trained on the feature representation. Since the regression layer in this case should not be included in training and might in this case lead to incomparable results.

\section{Discussion}

\begin{figure}[t]
	\begin{center}
		\includegraphics[width=0.93\linewidth]{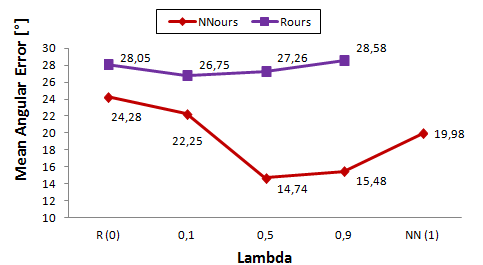}
		\vspace{0.1cm}
		\caption{Sensitivity of $\lambda$ in our loss function $L_{MTL} = (1 - \lambda) L_{pose} + \lambda L_{d}$. Depicted is the influence of different weighting parameters on the mean angular error for regression as well as nearest neighbor pose retrieval.} 
		\label{fig:lambda}
	\end{center}
\end{figure}

Using our proposed loss we were able to improve the pose regressions performance and robustness in respect to only using regression. However, the viewpoint descriptors learned with this loss seem to be much more discriminative compared to those learned with the triplet loss alone. Therefore, when we use them for the nearest neighbor search we get the best performance. Pose regression seems to be more difficult problem to solve alone, but when applied on a manifold learned features, where discrete poses are already well separated it becomes easier to perform pose regression and achieve better performance than by simple pose regression.
  
By including real data we were able to match synthetic and real images onto the same domain. Still it might not be possible to fully cover the pose space, which we found to significantly impact the pose regressions performance as some poses might not be mapped sufficiently. It is, however, still possible for the synthetic samples. In the ideal case, only rendered samples should be used in the training set, which would then also result in removing this limitation. This is one topic, that should be further researched in future work and which, we believe, can optimize our regression approach beyond the performance of nearest neighbor pose retrieval and create a generalized model, in turn optimizing the method's memory consumption and efficiency as well. 

Furthermore, the difference between the regression accuracy and the improved pose retrieval, based on nearest neighbor search, might also be due to the non-constrained space on which the poses are predicted. In our training set, we constrain the poses as well as in-plane rotations to a certain amount of degrees. However, while predicting the pose, no such constraint is enforced by the neural network. By constraining the network during training and enforcing geometric bounds, we should be able to improve the pose regression performance further. 

\raggedbottom
\section{Conclusion}

In this work, we have presented a multi-task learning framework for object recognition and pose estimation, which can be used for many applications, e.g. navigation in robotic grasping. By introducing a novel loss function, combining regression and manifold learning, we were able to improve both direct pose regression as well as nearest neighbor pose retrieval by a large margin, compared to the baseline methods. Thus, we conducted a detailed analysis of feature descriptor learning, regression and the effect that both tasks have on each other in the context of object pose estimation.


Future work includes improving our methods generalization capabilities by only using synthetic images for training, and removing the need to rely on real data at all.
\bibliographystyle{unsrt}

\end{document}